\documentclass[authoryear,preprint,12pt]{elsarticle}

\usepackage{amssymb}
\usepackage{amsmath}
\usepackage{enumitem}
\usepackage{makecell}
\usepackage{float}

\usepackage{tabularx}

\journal{Intelligent Systems with Applications}

\begin{document}

\begin{frontmatter}



\title{Quantum Computing Supported Adversarial Attack-Resilient Autonomous Vehicle Perception Module for Traffic Sign Classification}


\author[1]{Reek Majumder} %% Author name
\ead{rmajumd@g.clemson.edu}
\author[1]{Mashrur Chowdhury} 
\ead{mac@clemson.edu}
\author[2]{Sakib Mahmud Khan} %% Author name
\ead{sakibkhan@mitre.org}
\author[3]{Zadid Khan}
\ead{zadid.khan@walmart.com}
\author[4]{Fahim Ahmad}
\ead{ahmedf@email.sc.edu}
\author[5]{Frank Ngeni}
\ead{fngeni@scsu.edu}
\author[6]{Gurcan Comert}
\ead{gcomert@ncat.edu}
\author[5]{Judith Mwakalonge}
\ead{jmwakalo@scsu.edu}
\author[7]{Dimitra Michalaka}
\ead{dmichala@citadel.edu }

\affiliation[1]{organization={Glenn Department of Civil Engineering},%Department and Organization
            addressline={}, 
            city={Clemson},
            postcode={29631}, 
            state={SC},
            country={USA}}

\affiliation[2]{organization={MITRE Corporation },%Department and Organization
            addressline={}, 
            city={McLean},
            postcode={22102}, 
            state={VA},
            country={USA}}
\affiliation[3]{organization={Walmart Supply Chain (Transportation)},%Department and Organization
            addressline={}, 
            city={Bentonville},
            postcode={72716}, 
            state={AR},
            country={USA}}

\affiliation[4]{organization={Department of Civil and Environmental Engineering, University of South Carolina},%Department and Organization
            addressline={}, 
            city={Columbia},
            postcode={29208}, 
            state={SC},
            country={USA}}
\affiliation[5]{organization={Department of Engineering, South Carolina State University},%Department and Organization
            addressline={}, 
            city={Orangeburg},
            postcode={29117}, 
            state={SC},
            country={USA}}
\affiliation[6]{organization={Comp. Data Science and Engineering Department North Carolina A\&T State University},%Department and Organization
            addressline={}, 
            city={Greensboro},
            postcode={27411}, 
            state={NC},
            country={USA}}
\affiliation[7]{organization={Department of Civil and Environmental Engineering, The Citadel},%Department and Organization
            addressline={}, 
            city={Charleston},
            postcode={29409}, 
            state={SC},
            country={USA}}

%% Abstract
\begin{abstract}
Deep learning (DL)-based image classification models are essential for autonomous vehicle  (AV) perception modules since incorrect categorization might have severe repercussions. Adversarial attacks are widely studied cyberattacks that can lead DL models to predict inaccurate output, such as incorrectly classified traffic signs by the perception module of an autonomous vehicle. In this study, we create and compare hybrid classical-quantum deep learning (HCQ-DL) models with classical deep learning (C-DL) models to demonstrate robustness against adversarial attacks for perception modules. Before feeding them into the quantum system, we used transfer learning models, alexnet and vgg-16, as feature extractors. We tested over 1000 quantum circuits in our HCQ-DL models for projected gradient descent (PGD), fast gradient sign attack (FGSA), and gradient attack (GA), which are three well-known untargeted adversarial approaches. We evaluated the performance of all models during adversarial attacks and no-attack scenarios. Our HCQ-DL models maintain accuracy above 95\% during a no-attack scenario and above 91\% for GA and FGSA attacks, which is higher than C-DL models. During the PGD attack, our alexnet-based HCQ-DL model maintained an accuracy of 85\% compared to C-DL models that achieved accuracies below 21\%. Our results highlight that the HCQ-DL models provide improved accuracy for traffic sign classification under adversarial settings compared to their classical counterparts.
\end{abstract}

\begin{graphicalabstract}
Deep learning (DL)-based image classification models are essential for autonomous vehicle  (AV) perception modules since incorrect categorization might have severe repercussions. Adversarial attacks are widely studied cyberattacks that can lead DL models to predict inaccurate output, such as incorrectly classified traffic signs by the perception module of an autonomous vehicle. In this study, we create and compare hybrid classical-quantum deep learning (HCQ-DL) models with classical deep learning (C-DL) models to demonstrate robustness against adversarial attacks for perception modules. Before feeding them into the quantum system, we used transfer learning models, alexnet and vgg-16, as feature extractors. We tested over 1000 quantum circuits in our HCQ-DL models for projected gradient descent (PGD), fast gradient sign attack (FGSA), and gradient attack (GA), which are three well-known untargeted adversarial approaches. We evaluated the performance of all models during adversarial attacks and no-attack scenarios. Our HCQ-DL models maintain accuracy above 95\% during a no-attack scenario and above 91\% for GA and FGSA attacks, which is higher than C-DL models. During the PGD attack, our alexnet-based HCQ-DL model maintained an accuracy of 85\% compared to C-DL models that achieved accuracies below 21\%. Our results highlight that the HCQ-DL models provide improved accuracy for traffic sign classification under adversarial settings compared to their classical counterparts.
\end{graphicalabstract}

\begin{highlights}
\item Propose hybrid classical-quantum models to improve autonomous vehicle sign classification resilience.
\item Compare hybrid classical-quantum deep learning vs. classical deep learning models under gradient attack, fast gradient sign attack, and projected gradient sign attack.
\item Hybrid classical-quantum deep learning models maintain accuracy above 91\% under gradient attack and fast gradient sign attack.
\item AlexNet-based hybrid classical-quantum deep learning achieves 85\% accuracy under severe projected gradient sign attack vs. 21\% for its classical counterpart.

\item Use over 1000 quantum circuit combinations to evaluate the robustness of hybrid classical-quantum deep learning with alexnet and vgg-16 as feature extractors.
\end{highlights}

%% Keywords

\end{frontmatter}



\section{Introduction}
\label{Introduction}

\subsection{Background}
\label{Background}
Autonomous vehicles (AVs) widely use deep learning (DL) models to implement object detection \citep{Ren2017,Lin2017} and classification \citep{Liu2020} for their perception \citep{Pendleton2017} module tasks, such as detecting lanes, obstacles, and traffic signs. Nevertheless, these models are susceptible to adversarial attacks, and their performance deteriorates significantly when a carefully crafted perturbation/noise has been injected with the input image \citep{Szegedy2014,Moosavi-Dezfooli2016,Wei2019,Akhtar2018,Goodfellow2015}. Adversarial attacks that introduce small, nearly invisible changes to input images can manipulate the output from the DL models, thus proving disastrous to AV deployment. The severity of these adversarial attacks on DL models depends mainly on an attacker's goals and knowledge of the model. The attacker's intent can be to perform a targeted or non-targeted attack. In a targeted attack, the goal is to manipulate the model’s output to a particular output. In contrast, a non-targeted attack seeks to alter the model’s prediction to any incorrect output as long as it differs from the true output. These attacks have been categorized into three groups based on the attacker's knowledge of DL models: white-box, gray-box, and black-box attacks. An attacker that uses a white-box attack is aware of the trained DL model and creates carefully considered perturbations to the input data to deceive it. In a gray-box attack, an attacker knows the model’s design but is uninformed of its training weights. In a black-box attack, the attacker creates random disruption, with no knowledge of the model, to trick a trained model. 

Various defense strategies have been suggested, including image transformation or model re-training techniques. The image transformation technique involves input reconstruction \citep{Ren2020,Aprilpyone2019,Panda2019} and inputs denoising \citep{Ren2020} methods, which aim to pre-process the images before entering the DL models using techniques like smoothing, filtering (e.g., JPEG filter \citep{Liu2019,Das2017}, binary filter \citep{Xu2018}, random filtering \citep{Khan2022}),  and feature squeezing \citep{Xu2018}. Model re-training involves adversarial attack detection \citep{Feinman2017} and training \citep{Yuan2019} methods, where adversarial samples are generated using vigorous attacks, and the DL models are re-trained on adversarial samples. Another method called the defensive distillation technique \citep{Papernot2016} is another strategy that combines detection and training networks. In this method, the detection network creates the probability of vectors to label the original dataset, and the training network is used to re-train the model using the labeled adversarial samples dataset generated by the detection network. However, these techniques can perform well for known adversarial attacks used to create adversarial samples for re-training, but they can perform poorly for unknown attacks in the future; therefore, we need resilient DL models during adversarial attacks. Based on \citep{Björck2015} definition of cyber-resilience, we define DL resilience as the capability to correctly categorize the image, although malicious parties perturb the input image. 

Since quantum computing is becoming more mainstream, quantum computers have recently joined the race for high-performing computing systems due to their computational advantages of using quantum mechanical properties like superposition and entanglement. Theoretically, hilbert space for quantum systems \citep{Griﬃths2013} increases exponentially to system size, which makes it harder to simulate on classical computers. For example, a quantum system with tens and hundreds of qubits is classically intractable and proposed to demonstrate quantum supremacy over classical supercomputers \citep{Preskill2012}. Companies like Google have recently claimed quantum supremacy with its 53-qubit system named Sycamore, which will take classical supercomputers almost 10,000 years to solve \citep{Arute2019}. Furthermore, a 56-qubit H21 quantum computer from Quantinuum has surpassed Google's record by a factor of 100 \citep{Montanez-Barrera2025}.  

An experiment for classifying traffic signs with DL models during an adversarial attack in \citep{Papernot2017} incorrectly detected a stop sign as a speed sign which can lead to severe collisions if AVs operate with the help of these DL models.  In this regard, a possible approach to improving model robustness in AV applications is to use hybrid classical-quantum deep learning (HCQ-DL) models. With quantum layers within a DL, HCQ-DL structures can also take advantage of quantum entanglement and superposition to be more resistant to attacks by the adversary \citep{Majumdar2023,Baral2023}. The goal is to test the performance and resiliency of HCQ-DL models against adversarial attacks without using image transformation or model re-training techniques. We use transfer learning (TL) to extract features from pre-trained DL models like alexnet and vgg-16 before inputting the data to the quantum systems because currently available quantum processors, also known as noisy intermediate-scale quantum (NISQ) systems, cannot embed image data into a quantum system directly. These pre-trained models are developed using 1.2 million images for 1000 categories from the imagenet dataset \citep{Deng2009}. For image classification, the initial convolution layer of these models is frozen, which acts as a feature extractor. Finally, the last layers are replaced with custom layers of artificial neural networks (ANNs) or quantum neural networks (QNN) and tuned for our LISA traffic sign dataset \citep{LISADataset2009}. The LISA dataset consists of traffic signs taken from video shots from driving vehicles. We designed our QNN model consisting of low-depth variational quantum circuits (VQC) \citep{Endo2020,Mitarai2018,Benedetti2019,Khan2023}, which can be learned based on the quantum circuit learning framework \citep{Mitarai2018} for currently available NISQ hardware from IBM \citep{Wille2019}, Xanadu \citep{Bergholm2022} and Google \citep{Ho2018}.

\subsection{Contribution}
\label{Contribution}
To our knowledge, the performance and resilience comparison of the QML-based HCQ-DL with C-DL models has not been studied in the transportation sector. In this research, we tested over 1000 quantum circuit-based QNN layers for traffic sign classification by designing HCQ-DL models. We compared them against C-DL models for traffic sign classification, primarily for the AV sign perception module. Both our HCQ-DL and C-DL models are developed using alexnet and vgg-16 as feature extractors. Traditionally, vgg-16 models aimed to improve performance by using a deeper network with smaller filters (sixteen layers) than other models like alexnet (eight layers). C-DL models with alexnet and vgg-16 as feature extractors are vulnerable to adversarial attacks like gradient attacks (GA), fast gradient sign attacks (FGSA), and projected gradient descent attacks (PGD). Overall, our alexnet-based HCQ-DL model has a lower attack success rate of 0\%-8\% compared to its classical counterpart, with an attack success rate of 6\% to 73\%. Similarly, for vgg-based models, the HCQ-DL models show an attack success rate of 0\% to 77\%, while C-DL models range from 4\%-88\%. This study aims to address quantum-enabled robustness in AV perception, which presents opportunities for extending the security of AV systems against adversarial attacks using quantum-enabled DL architecture.

\subsection{Outline}
\label{Outline}

Section 2 discusses datasets describing the dataset's origin and attack dataset's generation. The creation of the C-DL and HCQ-DL models and a description of the performance metrics applied in our study are covered in Section 3, a discussion of the research method. Section 4 summarizes the findings of our investigation and shows how the model's performance changes as cyberattack intensity rises. The conclusion based on the findings is discussed in Section 5. Section 6 provides suggestions for additional analysis for future studies. 

\section{Dataset}
\label{Dataset}

This section introduces the development of a balanced LISA traffic sign dataset to train HCQ-DL and C-DL models for stop sign classification. Moreover, we also discuss the adversarial dataset generation for various white box attacks such as fast gradient sign attack (FGSA), gradient attack (GA), and projected gradient descent (PGD) attack to assess model robustness.

\subsection{Image Dataset}
\label{Image Dataset}
We used a portion of the extended LISA \citep{LISADataset2009} traffic sign dataset, which contains around 7,855 annotations from 6,610 video frames identifying 47 different traffic signs. To examine the performance of the HCQ-DL and C-DL models, we focused on stop signs and a combination of other signs to design a balanced dataset. Image frames vary from 640 x 480 to 1024 x 522 pixels. The size of annotation boxes for traffic signs ranges from 6 x 6 to 167 x 168 pixels. The number of samples for each type of traffic sign differs significantly. Initially, we grouped the traffic sign dataset into 18 traffic signs and cropped up the images to reduce the noise in their surroundings.
\begin{figure}[htbp]
    \centering
    \includegraphics[width=0.8\textwidth]{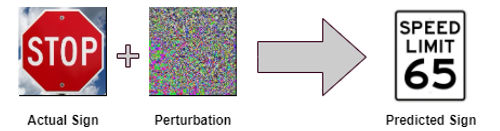}
    \caption{Example of adversarial attack on stop signs.}
    \label{fig:overview}
\end{figure}

In the binary classification models, both HCQ-DL and C-DL models are trained to classify stop signs and other signs by creating a balanced dataset. On the extracted balanced dataset, which included 231 samples with an 80-20 split between training and testing data, both HCQ-DL and C-DL models were trained and tested for twenty-five epochs (number of training samples:182, number of testing samples: 49). 

\subsection{Attack Models for Adversarial Dataset}
\label{Attack Models for Adversarial Dataset}
In this study, three types of white-box attacks have been chosen based on severity (elementary, intermediate, and advanced), and attacks generated with fast gradient sign attack (FGSA) \citep{Goodfellow2015}, gradient attack (GA) \citep{Goodfellow2015}, and projected gradient descent (PGD) attack \citep{Madry2017} for both HCQ-DL and C-DL models. Before the image is entered into the classification model, these attacks create perturbations to the input data based on the epsilon coefficient. These attacks differ in how the perturbation is applied to the original input images to generate misclassified outputs. Equation 1 illustrates how the gradient attack modifies the input image considering the gradient of the loss function for the DL model to produce an adversarial image. Moreover, the FGSA is a single-step gradient ascent strategy that uses a sign of gradient with a fixed epsilon coefficient to generate an adversarial image, as shown in equation 2. However, PGD obtains adversarial samples by iteratively using the fast gradient method, and the iteration starts uniformly at a randomly chosen data point. It projects (Proj) the adversarial samples from each iteration into the next using the product of epsilon and the gradient of loss function, as represented in equation 3 \citep{Khan2022}.

\begin{align}
\text{Adversarial image} &= \text{input image} + \left[ \epsilon \cdot \nabla_{\text{input image}} J \right] \tag{1} \\
\text{Adversarial image} &= \text{input image} + \left[ \epsilon \cdot \text{sign} \left( \nabla_{\text{input image}} J \right) \right] \tag{2} \\
% \text{Iterative Adversarial Image} &= \text{Projection} \big( \text{Adversarial image}_{\text{prev}} + 
% \epsilon \cdot \text{sign} \left( \nabla_{\text{input image}} J \right) \big) \tag{3}
\text{Iterative Adversarial Image} &= 
\text{Projection} \Big( \text{Adversarial image}_{\text{prev}} \nonumber \\
&\quad + \epsilon \cdot \text{sign} \left( \nabla_{\text{input image}} J \right) \Big) \tag{3}
\end{align}

All attacks studied during this experiment modify input images to deceive DL models, as shown in Figure 1, and attack intensity has varied from perturbation (epsilon) coefficients ranging from 0.05 to 0.5, discussed in Section IV.

\section{Research Method}
\label{Research Method}
\begin{figure}[htbp]
    \centering
    \includegraphics[width=0.8\textwidth]{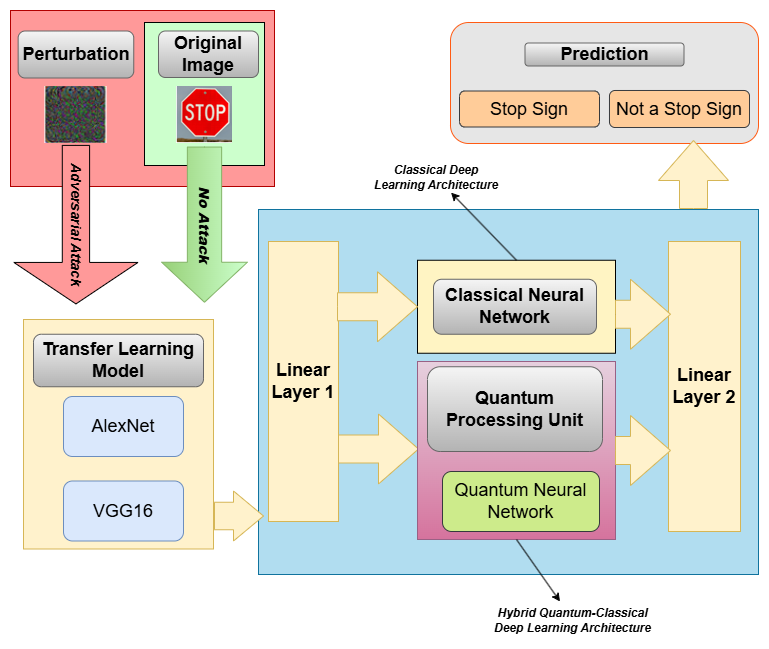}
    \caption{Architecture for C-DL and HCQ-DL models.}
    \label{fig:overview}
\end{figure}
\subsection{Transfer Learning}
A well-known machine learning method for feature extraction and handling situations when we lack training data is transfer learning (TL) \citep{Majumdar2023,Zhuang2021,Mari2020}. Since we have only 182 samples in our training set and want to encode image data to a quantum system, we use known TL models as feature extractors (alexnet and vgg-16). For image classification tasks, initial convolution layers of these TL models are believed to learn similar features, so their weights are fixed and extended to fine-tune on newer tasks by replacing final layers with ANNs.

These TL models are trained on the imagenet dataset with 1.2 million images for over 1000 categories. The objective behind employing TL models is to utilize the initial convolution layer of these pre-trained models and swap out the final linear layer with layers of custom linear layers according to the given use case. 

% \begin{figure}[htbp]
%     \centering
%     \includegraphics[width=0.8\textwidth]{Fig2.png}
%     \caption{Example of adversarial attack on stop signs.}
%     \label{fig:overview}
% \end{figure}
% \clearpage

\subsection{Classical Deep Learning (C-DL) and Hybrid Classical-Quantum Deep Learning (HCQ-DL) Models}
\label{C_DL_and_HCQ_DL}
This section describes the development of the C-DL and HCQ-DL model architecture used in our study for training and testing traffic sign classifiers. We also represent the final hyperparameters of our HCQ-DL and C-DL models with vgg-16 and alexnet pre-trained models as feature extractors.

% \clearpage

\subsubsection{Classical Deep Learning (C-DL) Models }

As shown in figure 2, our experiment uses two state-of-the-art convolutional neural network-based deep learning models, vgg-16 \citep{Simonyan2014} and alexnet \citep{Krizhevsky2012}, as our feature extractors for our C-DL and HCQ-DL models. Due to the lack of training samples, we use transfer learning principles for the image classification task and freeze the initial layers of the pre-trained model. Later we introduced two linear layers for our C-DL model (figure 2) with a rectilinear unit (ReLU) as an activation function for linear layer one and SoftMax for the final linear layer. We later trained these C-DL models on stop signs and other sign classes. 

\subsubsection{Hybrid Classical-Quantum Deep Learning (HCQ-DL) Models}

We developed and tested over 1000 quantum circuits-based QNN models with pre-trained DL- models to develop our HCQ-DL models. The QNN models were composed of various single qubits and multiple qubits gates. These gates are fundamental building blocks for circuit-based quantum algorithms. Usually, these gates are evaluated based on the impact of each gate on 3-dimensional space for each qubit, often referred to as the bloch sphere. The gates used in our study are explained below. 

\begin{enumerate}[label=\textbf{\alph*.}, leftmargin=2em]

\item \textbf{Hadamard gate:} A fundamental quantum gate that helps create a superposition of states \(|0⟩\) and \(|1⟩\) by moving away from the poles of the Bloch sphere\citep{Simonyan2014}.

\item \textbf{Rotational gates:} These include three Pauli rotational gates — rotational-x (RX), rotational-y (RY), and rotational-z (RZ). These gates rotate the state vector about the corresponding axis and are often generated by exponentiating the Pauli operators. Specifically, Pauli-X (X) is a bit-flip gate, Pauli-Y (Y) is a bit- and phase-flip gate, and Pauli-Z (Z) is a phase-flip gate\citep{Crooks2020}.

\item \textbf{Universal gates:} These advanced single-qubit gates combine rotational and phase-shift operations to represent different states on the Bloch sphere. The most common are:
\begin{itemize}
    \item U1: equivalent to a phase shift gate,
    \item U2: combines Y- and Z-axis rotations with a phase shift,
    \item U3: applies rotation about all three Bloch sphere axes\citep{Barenco1995,Author46n.d.}.
\end{itemize}

\item \textbf{Controlled gates:} Our study employs 2-qubit controlled gates (control and target) using Pauli operators (X, Y, Z) and rotational gates (RX, RY, RZ). These gates act on the target qubit only when the control qubit is in the \(|1⟩\) state. This selective behavior introduces entanglement — a correlated state — among qubits, which is one of the unique features of our approach.

\item \textbf{Measurement gate:} All 1000 quantum circuits include a final measurement gate that maps quantum states to classical bits. Each circuit was run 1000 times per sample to obtain a probability distribution over outcomes. The state with the highest probability was used as input to the next phase, namely linear layer 2 (see Figure~2). Figure~3 illustrates the QNN architecture and the optimal quantum circuit in our HCQ-DL model, where VGG-16 and AlexNet serve as classical feature extractors. As quantum computers cannot directly process images \citep{Baral2023,Mitarai2018,Schuld2018}, the HCQ-DL framework first preprocesses images using pretrained CNNs, then replaces the final layer with a quantum layer sandwiched between two linear layers, as depicted in Figure~2.

\end{enumerate}

\begin{table}[t]
\centering
\caption{\textbf{Model Hyperparameters and Training Time}}
\label{tab:model_params}
\vspace{1em}
\renewcommand{\arraystretch}{1.5}
\begin{tabular}{|>{\centering\arraybackslash}p{3.9cm}|
                >{\centering\arraybackslash}p{1.9cm}|
                >{\centering\arraybackslash}p{1.9cm}|
                >{\centering\arraybackslash}p{1.9cm}|
                >{\centering\arraybackslash}p{1.9cm}|}
\hline
\makecell{ \textbf{Transfer Learning } \\ \textbf{Model}} & 
\makecell{\textbf{VGG-16} \\ (Classical)} & 
\makecell{\textbf{AlexNet} \\ (Classical)} & 
\makecell{\textbf{VGG-16} \\ (Hybrid)} & 
\makecell{\textbf{AlexNet} \\ (Hybrid)} \\
\hline
\textbf{Scheduler Step Size} & 8 & 9 & 8 & 9 \\
\textbf{Batch Size} & 32 & 64 & 2 & 8 \\
\textbf{Learning Rate} & 0.000697 & 0.000269 & 0.000194 & 0.00291 \\
\textbf{Optimizer} & Adam & Adam & Adam & Adam \\
\makecell{\textbf{Classical (Neuron)}/ \\ \textbf{Hybrid (N-Qubits)}} & \makecell{\\ \\ 2} & \makecell{\\ \\ 6} & \makecell{\\ \\ 4} & \makecell{\\ \\ 3} \\
\textbf{Training Time} & 
14 min 11 sec & 
2 min 27 sec & 
47 min 38 sec & 
28 min 11 sec \\
\makecell{\textbf{Number of} \\ \textbf{Parameters}} & 
14,764,872 & 
2,525,012 & 
14,815,066 & 
2,497,370 \\
\hline
\end{tabular}
\end{table}

Furthermore, we divide the QNN layer into four broad categories. The data encoding layer embeds data from the classical system into the quantum system. The variational quantum circuit combines repetitive single and multi-qubit gates, a pre-measurement layer consisting of single-qubit gates, and a measurement layer that reads data from the quantum system to the classical system. We have optimized the first linear layer's number of neurons, ranging from 2 to 8, for the C-DL, and it refers to the number of qubits needed to initialize for quantum layers of HCQ-DL models. The ranges of other DL parameters are scheduled step sizes ranging from 5 to 10 and learning rates ranging from 0.01 to 0.0001. We consider batch sizes 2, 4, 8, 16, 32, and 64. We tested our models for adam and stochastic gradient descent (SGD) optimizer and reported the tuned model parameters for our best models in Table I.
\begin{figure}[htbp]
    \centering
    \includegraphics[width=0.75\textwidth]{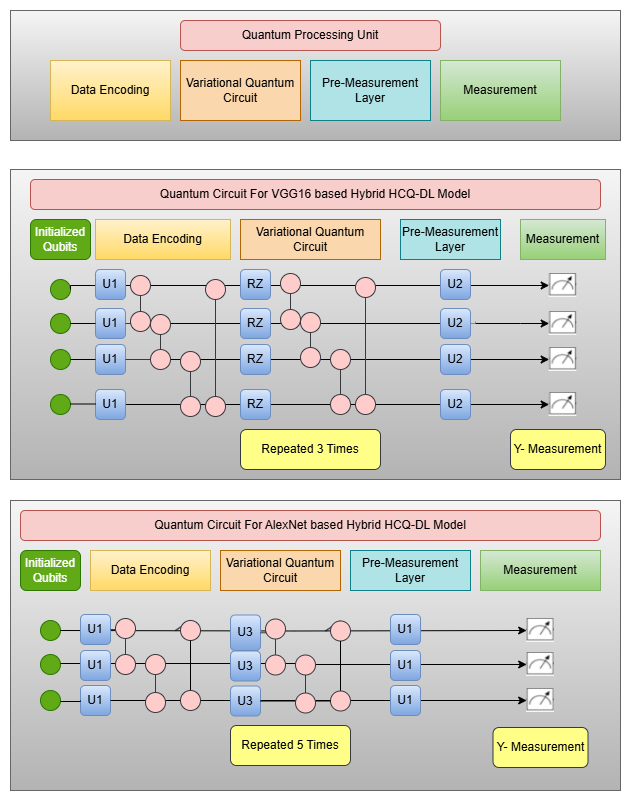}
    \caption{Quantum Circuit Architecture with chosen Circuits for our VGG16-based HCQ-DL and AlexNet-based HCQ-DL models with better resiliency.}
    \label{fig:overview}
\end{figure}

Comparing hyperparameters of alexnet vs. vgg-16 C-DL models in Table 1, we found that alexnet models have relatively lower training time due to fewer parameters. Lower training time due to a lower number of parameters can be seen as consistent in the HCQ-DL alexnet model vs. the HCQ-DL vgg-16 model. However, HCQ-DL models have considerably higher training time even after having fewer parameters because we use quantum simulators from pennyLane \citep{Benedetti2019} and cannot exclude the wait time for each batch during training. With the improvement in performance and availability of quantum hardware, the computation time will likely go down. 

\subsection{Quantum Gradient Calculation and Backpropagation in HCQ-DL Models}
Gradient-based optimization for HCQ-DL models combines classical backpropagation methods with quantum differentiation techniques. The pre-trained CNN model (vgg-16 or alexnet) performs image feature extraction during the forward pass. The data encoding layer of QNN transforms the extracted features into a quantum representation. The quantum representation performs data processing through VQCs using trainable unitary transformation before output probabilities are extracted for classification using the measurement layer.

The backward pass computes gradients for both the classical neural network and quantum circuit layers. Gradients of pre-trained CNN layers are kept constant because their weights have been frozen. The final linear layers calculate their gradients using traditional backpropagation methods. Quantum circuits differ from classical networks because they lack analytical derivatives, which require an alternate approach to calculate quantum gradients. We employ the parameter-shift rule that numerically computes gradients through small adjustments in the quantum parameters, as shown in equation 4.

\begin{equation}
\frac{\partial f(\theta)}{\partial \theta} = \frac{f(\theta + s) - f(\theta - s)}{2s}
\tag{4}
\end{equation}

\noindent where $s$ is a small shift. This technique ensures that the QNN layer can be trained end-to-end along with the classical layers.

\subsection{Performance Matrix}

The C-DL and HCQ-DL models are evaluated using accuracy for training and testing, where accuracy is defined as the proportion of correct predictions to all predictions for samples in the train and test set. For our investigation of adversarial attacks, we selected the models with higher accuracy.  

We tested our models with other performance metrics like sensitivity, specificity, false positive rate, precision, and F1 score. Specificity refers to the proportion of other sign classes that are correctly classified. At the same time, sensitivity, also known as recall, relates to the fraction of stop sign classes that are correctly classified. We also calculated the false positive rate, which refers to the fraction of other sign class samples misclassified as a stop sign. We also calculated the precision for each model to study how well the models are classifying stop signs and other sign classes and tested the balance between precision and recall by evaluating the F1 score of our model. 

\begin{equation}
MCC = 
\left( \frac{(TP \cdot TN) - (FP \cdot FN)}{\sqrt{(TP + FP)(TN + FN)}} \right) 
\cdot 
\left( \frac{1}{\sqrt{(TP + FN)(TN + FP)}} \right)
\tag{5}
\end{equation}

\noindent\textit{Where TP: True Positive, FP: False Positive, TN: True Negative, FN: False Negative}

Finally, we tested C-DL and HCQ-DL-based classifier models on statistical measure phi-coefficient, also referred to as matthew's correlation coefficient (MCC), which measures the correlation between the actual classes of the dataset and the predicted classes by the models and calculated as shown in equation 5.

\section{Analysis}

\begin{figure}[htbp]
    \centering
    \includegraphics[width=0.77\textwidth]{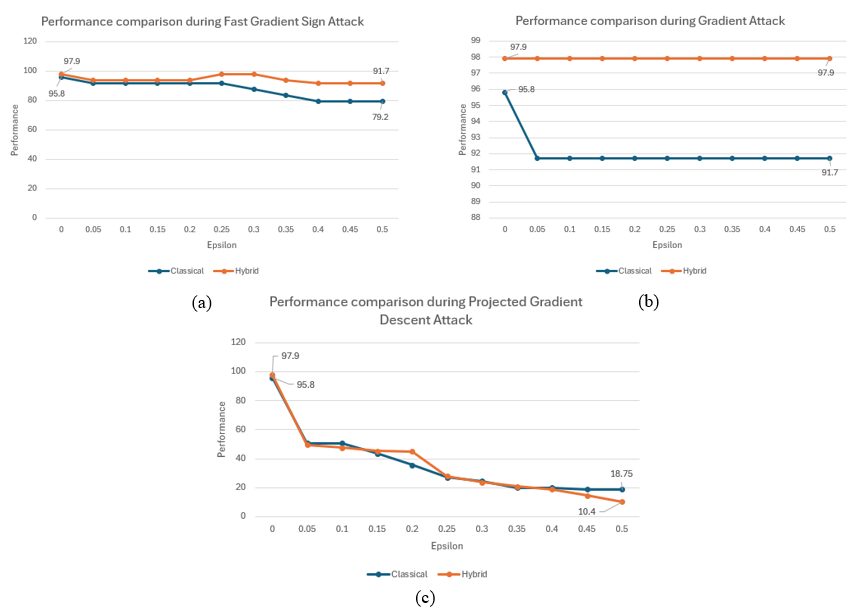}
    \caption{Performance Comparison of VGG16-based C-DL and HCQ-DL models under varying intensity of perturbation coefficients for: a) Fast Gradient Sign Attack b) Gradient Attack c) Projected Gradient Descent Attack.}
    \label{fig:overview}
\end{figure}

\begin{figure}[htbp]
    \centering
    \includegraphics[width=0.77\textwidth]{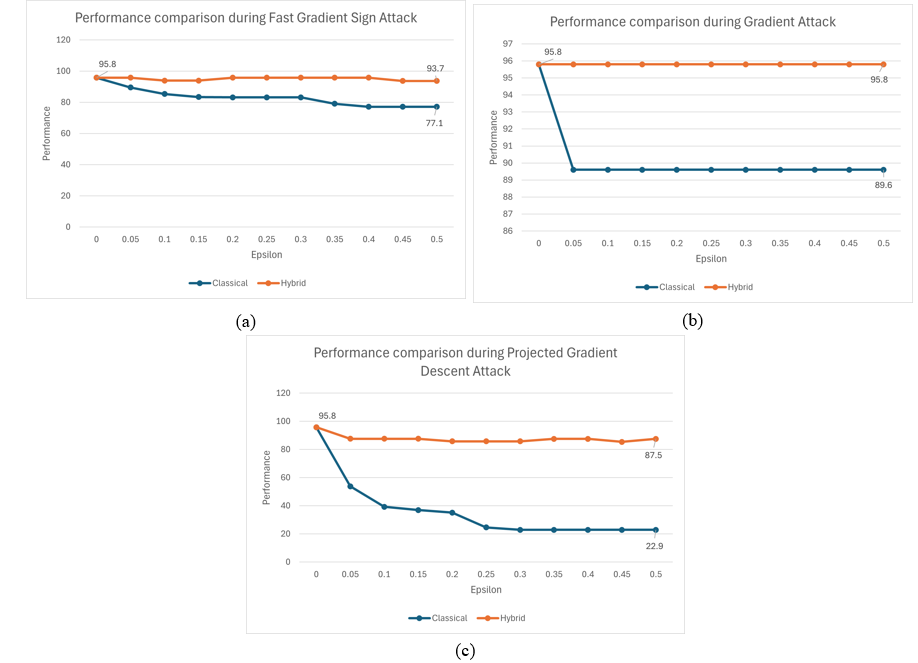}
    \caption{Performance Comparison of Alexnet-based C-DL and HCQ-DL models under varying intensity of perturbation coefficients for: a) Fast Gradient Sign Attack b) Gradient Attack c) Projected Gradient Descent Attack.}
    \label{fig:overview}
\end{figure}

\begin{table}[t]
\centering
\caption{\textbf{Model Performance Metrics}}
\vspace{1em}
\label{tab:model_performance}
\renewcommand{\arraystretch}{1.5}
\begin{tabular}{|>{\centering\arraybackslash}p{3.7cm}|
                >{\centering\arraybackslash}p{1.8cm}|
                >{\centering\arraybackslash}p{1.8cm}|
                >{\centering\arraybackslash}p{1.7cm}|
                >{\centering\arraybackslash}p{1.7cm}|}

% {|p{4cm}|p{2cm}|p{2cm}|p{2cm}|p{2cm}|}
\hline
\makecell{\textbf{Transfer Learning } \\ \textbf{Model}} & 
\makecell{\textbf{VGG-16} \\ (Classical)} & 
\makecell{\textbf{AlexNet} \\ (Classical)} & 
\makecell{\textbf{VGG-16} \\ (Hybrid)} & 
\makecell{\textbf{AlexNet} \\ (Hybrid)} \\
\hline
\makecell{\textbf{Accuracy}} & 96\% & 96\% & 98\% & 96\% \\
\makecell{\textbf{Specificity}} & 1 & 1 & 1 & 1 \\
\makecell{\textbf{Sensitivity}} & 0.90 & 0.90 & 0.95 & 0.90 \\
\makecell{\textbf{False Positive} \\ \textbf{Rate}} & 0.10 & 0.10 & 0.05 & 0.10 \\
\makecell{\textbf{Precision Score} } & 0.93 & 0.93 & 0.97 & 0.93 \\
\makecell{\textbf{F1-Score}} & 0.97 & 0.97 & 0.98 & 0.97 \\
\makecell{\textbf{MCC Score} } & 0.92 & 0.92 & 0.96 & 0.92 \\
% \makecell{\textbf{Matthews Correlation } \\ \textbf{Coefficient (MCC)}} & 
% \makecell{0.92} & 
% \makecell{0.92} & 
% \makecell{0.96} & 
% \makecell{0.92} \\
\hline
\end{tabular}
\end{table}
This section discusses the performance results for our C-DL and HCQ-DL models. And QNN models architecture for best HCQ-DL models. Our best vgg-16-based HCQ-DL model, we initialize a 4-qubit system, where during the data encoding phase, we use a U1 gate with a controlled-Z (CZ) entanglement gate, while in a variational quantum circuit includes RZ and CZ gate repeated for three times and U2 gate in pre-measurement. In the alexnet-based HCQ-DL model, we initialize a 3-qubit system, where the data encoding phase uses a U1 gate with a CZ entanglement gate, while our variational quantum circuit includes U3, and CZ gates repeated five times and a U1 gate in pre-measurement layer. Finally, we perform quantum Y-measurement for both models to map quantum states to classical output. The circuits are executed for 1000 shots, and the state with maximum probability is considered as quantum measurement output. The measurement output is mapped to the ANN layer with neurons equivalent to the qubit system initialized to each model, as shown in Figure 2. Finally, an ANN layer of two neurons with a softmax activation function provides a final output for all our models.

% \begin{table}[t]
% \centering
% \caption{\textbf{Model Performance Metrics}}
% \label{tab:model_performance}
% \renewcommand{\arraystretch}{1.3}
% \begin{tabular}{|l|c|c|c|c|}
% \hline
% \textbf{Transfer Learning Model} & \multicolumn{2}{c|}{\textbf{Classical Model}} & \multicolumn{2}{c|}{\textbf{Hybrid Classical-Quantum Model}} \\
% \hline
% & \textbf{VGG-16} & \textbf{AlexNet} & \textbf{VGG-16} & \textbf{AlexNet} \\
% \hline
% \textbf{Accuracy} & 96\% & 96\% & 98\% & 96\% \\
% \textbf{Specificity} & 1 & 1 & 1 & 1 \\
% \textbf{Sensitivity} & 0.90 & 0.90 & 0.95 & 0.90 \\
% \textbf{False Positive Rate} & 0.10 & 0.10 & 0.05 & 0.10 \\
% \textbf{Precision Score} & 0.93 & 0.93 & 0.97 & 0.93 \\
% \textbf{F1-Score} & 0.97 & 0.97 & 0.98 & 0.97 \\
% \textbf{Mathew Correlation} & & & & \\
% \textbf{Coefficient (MCC)} & 0.92 & 0.92 & 0.96 & 0.92 \\
% \hline
% \end{tabular}
% \end{table}

\begin{table}[t]
\centering
\caption{\textbf{Perturbation Coefficient (PC) at which Models Record its Lowest Accuracy}}
\label{tab:perturbation_pc}
\vspace{1em}
\renewcommand{\arraystretch}{1.5}
\begin{tabular}{|>{\centering\arraybackslash}p{4cm}|
                >{\centering\arraybackslash}p{2.1cm}|
                >{\centering\arraybackslash}p{1.9cm}|
                >{\centering\arraybackslash}p{1.9cm}|
                >{\centering\arraybackslash}p{1.8cm}|}

% {|p{4.5cm}|p{2cm}|p{2cm}|p{2cm}|p{2cm}|}
\hline
\textbf{Transfer Learning Model} & 
\makecell{\textbf{VGG-16} \\ (Classical)} & 
\makecell{\textbf{AlexNet} \\ (Classical)} & 
\makecell{\textbf{VGG-16} \\ (Hybrid)} & 
\makecell{\textbf{AlexNet} \\ (Hybrid)} \\
\hline
\makecell{\textbf{Accuracy} \\ (PC)} & 96\% (0.0) & 96\% (0.0) & 98\% (0.0) & 96\% (0.0) \\
\makecell{\textbf{Gradient} \\ \textbf{Attack (PC)}} & 92\% (0.05) & 90\% (0.05) & 98\% (0.5) & 96\% (0.5) \\
\makecell{\textbf{Fast Gradient} \\ \textbf{Sign (PC)}} & 79\% (0.35) & 77\% (0.4) & 92\% (0.4) & 94\% (0.45) \\
\makecell{\textbf{Projected Gradient} \\ \textbf{Descent Attack} \\ \textbf{(PC)}} & \makecell{\\ \textbf{19\% (0.45)}} & \makecell{\\ \textbf{23\% (0.3)}} & \makecell{\\ \textbf{10\% (0.5)}} & \makecell{ \\ \textbf{85\% (0.2)}} \\
\hline
\end{tabular}
\end{table}

Table 2 represents the complete report regarding the accuracy, sensitivity, specificity, false positive rate, precision, F1 score, and MCC for each of our best C-DL and HCQ-DL models. We analyzed these parameters and found that all our models satisfy the criteria of higher sensitivity, specificity, precision, F1-score, and MCC while having a lower false positive rate, which is an essential criterion for a good classifier. As a result of our analysis of these variables, we discovered that all our models meet the crucial criteria of a strong classifier: higher sensitivity, specificity, accuracy, precision F1-score, and MCC while having lower false positive rates.

Figures 4 and 5 show the performance change for alexnet-based and vgg16-based HCQ-DL and C-DL models during GA, FGSA, and PGD attacks with a perturbation coefficient ranging from 0.05 to 0.5 with an interval of 0.05. We can see from figures 4 and 5 that vgg-16 and alexnet-based HCQ-DL models are more resilient and maintain higher accuracy even after increasing the intensity of GA and FGSA attacks. 

The most aggressive attack considered in our analysis is projected gradient descent (PGD), and figure 18c and 19c shows the change in performance of our model with increasing intensity of PGD attack. We can see that the AlexNet-based HCQ-DL model exponentially outperforms all other models by having a lower attack success rate and maintaining higher accuracy. The worst results of our C-DL and HCQ-DL models are shown in Table 3 during adversarial attacks with the perturbation coefficient. 
% \begin{table}[t]
% \centering
% \caption{\textbf{Perturbation Coefficient (PC) at which Models Record its Lowest Accuracy}}
% \label{tab:perturbation_pc}
% \renewcommand{\arraystretch}{1.3}
% \begin{tabular} {|p{3cm}|p{1.5cm}|p{1.5cm}|p{1.5cm}|p{1.5cm}|}
% % {|l|c|c|c|c|}
% \hline
% \textbf{Transfer Learning Model} & \multicolumn{2}{c|}{\textbf{Classical Model}} & \multicolumn{2}{c|}{\textbf{Hybrid Classical-Quantum Model}} \\
% \hline
% & \textbf{VGG-16} & \textbf{AlexNet} & \textbf{VGG-16} & \textbf{AlexNet} \\
% \hline
% \textbf{Accuracy (PC)} & 96\% (0.0) & 96\% (0.0) & 98\% (0.0) & 96\% (0.0) \\
% \textbf{Gradient Attack (PC)} & 92\% (0.05) & 90\% (0.05) & 98\% (0.5) & 96\% (0.5) \\
% \textbf{Fast Gradient Sign (PC)} & 79\% (0.35) & 77\% (0.4) & 92\% (0.4) & 94\% (0.45) \\
% \textbf{Projected Gradient} & \textbf{19\% (0.45)} & \textbf{23\% (0.3)} & \textbf{10\% (0.5)} & \textbf{85\% (0.2)} \\
% \textbf{Descent Attack (PC)} & & & & \\
% \hline
% \end{tabular}
% \end{table}

\section{Conclusion}

From alexnet to vgg-16, CNN-based DL models have historically evolved, intending to add more convolutional layers for better data mapping. In this investigation, we discovered that the performance and resiliency of DL models to adversarial attacks could be enhanced using HCQ-DL models built by combining quantum layers with the current C-DL model. In our study, we obtained better performance accuracy in recognizing the sign under adversarial attacks, establishing the resiliency while not using well-known defenses such as image modification or model re-training on adversarial samples, which would take higher processing power and time. Image modification techniques are pre-processing steps that emphasize smoothing, while filtering introduces additional implementation steps during the training and testing phase without considering the type of attack.  While model re-training methods emphasize re-training on adversarial images, they do not account for unknown adversarial attacks, which could be more malicious than the known attack models based on which these are re-trained. Our HCQ-DL method presented in this study offers a performance accuracy boost during a known or unknown adversarial attack by introducing a quantum network within C-DL models without requiring pre-processing procedures like image modification or post-processing procedures like model-re-training. This occurs because quantum systems are known for mapping various counter-intuitive patterns, which is leveraged in our HCQ-DL model development.

Higher accuracy, precision, recall, F1 score, specificity, mathew's correlation coefficient, and a lower false positive rate are typically required for a classifier to function well. Our study shows that both our C-DL and HCQ-DL satisfy these criteria; however, they are vulnerable to adversarial attacks. Our study showed that compared to C-DL models, HCQ-DL models maintain a higher accuracy (above 95\%) during gradient attacks and above 90\% during Fast Gradient sign attacks. For the projected gradient descent attack, we found that the alexnet-based hybrid model outperforms other HCQ-DL and C-DL models by constantly having an accuracy above 85\%.

Our alexnet-based HCQ-DL shows better performance and resiliency during adversarial attacks than vgg-16-based HCQ-DL models. This opens the possibility of looking into shorter networks with quantum layers, resulting in fewer parameters but still maintaining a higher level of feature mapping, thus improving the performance accuracy and resilience of these next-generation models against adversarial attacks.

\section{Future Work}

Our research demonstrates that the C-DL architecture may be considerably strengthened by adding a single quantum layer to increase the robustness of deep learning models against adversarial attacks. HCQ-DL models can maintain relatively higher accuracy for highly effective adversarial attacks like L-infinite projected gradient descent attacks. 

In our future work, we will evaluate the effect of the perturbation on the internal layers of DL models, which are used as feature extractors for our HCQ-DL models. We also intend to test these models for different lighting and environmental conditions to evaluate their robustness in real-world AV deployments. We have used error-free quantum simulators from pennylane for this study. In the future, we would also like to train and test these models on actual quantum computers and address quantum errors and noise, usually present in physical quantum computers. 

Furthermore, our future work will investigate Lipschitz-based regularization implementation in quantum layers, specifically for the encoding layer to regulate model sensitivity to adversarial perturbations beyond the current outlined directions.  Studies show that this method decreases vulnerability while simultaneously enhancing the model's focus on significant features in input data to boost both interpretability and robustness. We will investigate the impact of regularization on model generalization and resilience against unknown and transferable attacks through theoretical analysis using Lipschitz bounds and practical evaluation of attack transfer performance as suggested in \citep{Wendlinger2024}. 

Finally, we intend to evaluate how adversarial attacks transfer between classical and quantum domains. Some recent research suggests that Fourier-based classical approximation of quantum models can provide an intermediary framework for analyzing robustness at the classical-quantum boundary. Investigating this area allows us to analyze how hybrid models respond to adversarial attacks and to discover design principles for creating robust architectures.

\section{Acknowledgement}

This work was partially supported by the Centre for Connected Multimodal Mobility(C2M2) (the US Department of Transportation Tier 1 University Transportation Centre) and the National Center for Transportation Cybersecurity and Resiliency (TraCR) (a U.S. Department of Transportation National University Transportation Center), headquartered at Clemson University, Clemson, SC, USA. Any opinions, findings, conclusions, and recommendations expressed in this material are those of the author(s) and do not necessarily reflect the views of C2M2 and TraCR, and the US government assumes no liability for the contents and use thereof.

During the preparation of this work, the authors used ChatGPT in order to improve the readability and language of the work. After using chatGPT, the author(s) reviewed and edited the content as needed and takes full responsibility for the content of the publication.

\section{Declaration}

\subsection{Ethical Approval}

It did not require any ethical approval. There was no human and animal study in the research presented in this paper.

\subsection{Competing interests }
There is no competing interest that we are aware off.
\begin{table}[h!]

\centering
\caption{Author Contributions}
\vspace{1em}
\begin{tabular}{|l|p{9cm}|}
\hline
\textbf{Name} & \textbf{Contribution} \\
\hline
Reek Majumder & Conceptualization, Formal Analysis, Investigation, Methodology, Validation, Visualization, Writing - original draft \\
\hline
Mashrur Chowdhury & Conceptualization,  Writing - original draft, Funding acquisition \\
\hline
Sakib Mahmud Khan &  Conceptualization, Writing - review and editing, Supervision \\
\hline
Zadid Khan & Methodology, Writing - review and editing \\
\hline
Fahim Ahmed & Validation, Writing - review and editing \\
\hline
Frank Ngeni & Validation, Writing - review and editing \\
\hline
Gurcan Comert & Visualization, Writing - review and editing \\
\hline
Judith Mwakalonge & Writing - review and editing \\
\hline
Dimitra Michalaka & Writing - review and editing \\
\hline
\end{tabular}
\label{tab:author_contributions}
\end{table}

\subsection{Authors' contributions }
Table 4 shares the author contribution involved in this paper

\subsection{Funding}
was partially supported by the Centre for Connected Multimodal Mobility(C2M2) (the US Department of Transportation Tier 1 University Transportation Centre) and the National Center for Transportation Cybersecurity and Resiliency (TraCR) (a U.S. Department of Transportation National University Transportation Center), headquartered at Clemson University, Clemson, SC, USA. Any opinions, findings, conclusions, and recommendations expressed in this material are those of the author(s) and do not necessarily reflect the views of C2M2 and TraCR, and the US Government assumes no liability for the contents and use thereof.
\subsection{Availability of data and materials}
The dataset was built using the LISA traffic sign dataset containing 47 different signs. We grouped the signs into 18 traffic signs and cropped the images to reduce the noise in their surroundings. Finally, for this study, we created a balanced dataset with stop and other signs. 
The dataset used, and codes developed for this study can be found in the GitHub  repository: https://github.com/reek129/QuantumAI\_MultiClass\_Sign\_Recognition

\bibliographystyle{elsarticle-harv}

\end{document}